\def\mycmd{2}

\if\mycmd1
\documentclass[11pt,onecolumn, draftcls]{IEEEtran}
\else 
\documentclass[lettersize,journal]{IEEEtran}
\fi
\usepackage[utf8]{inputenc}
\usepackage{cite}
\usepackage{stfloats}
\usepackage{enumitem}
\usepackage{kotex}
\usepackage{amsmath,amssymb,amsfonts}
\usepackage{algorithmic, algorithm}
\usepackage[dvipsnames]{xcolor}
\usepackage{graphicx}
\usepackage{mathtools, empheq} 
\usepackage{textcomp}
\usepackage{xcolor}
\usepackage{booktabs}
\usepackage{subcaption}
\usepackage{array}

\def\BibTeX{{\rm B\kern-.05em{\sc i\kern-.025em b}\kern-.08em
    T\kern-.1667em\lower.7ex\hbox{E}\kern-.125emX}}
    
\begin{document}
\title{Collaborative Large Language Model Inference via
Resource-Aware Parallel Speculative Decoding}
\author{Jungyeon Koh,~\IEEEmembership{Member,~IEEE}, and Hyun Jong Yang,~\IEEEmembership{Senior Member,~IEEE}
\thanks{Jungyeon Koh is with the Department of Electrical Engineering, Pohang University of Science and Technology (POSTECH), Korea (e-mail: jungyeon.koh@postech.ac.kr). Hyun Jong Yang is with the Department of Electrical and Computer Engineering, Seoul National University, Korea (email: hjyang@snu.ac.kr).
}
}

\maketitle
\begin{abstract}\label{abstract}
The growing demand for on-device large language model (LLM) inference highlights the need for efficient mobile edge computing (MEC) solutions, especially in resource-constrained settings. Speculative decoding offers a promising solution by partitioning token generation between a lightweight draft model on mobile devices and a powerful target model on edge servers, but suffers from communication overhead and asynchronous delays. This paper propose a unified framework that jointly optimizes user association and resource allocation (UARA) to support efficient parallel speculative decoding. We solve the UARA problem using a multi-agent deep reinforcement learning algorithm. Simulation results show that our method achieves up to 28.0\% and an average of 23.7\% reduction in end-to-end latency without compromising inference accuracy, enabling scalable and low-latency LLM services in MEC systems.
\end{abstract}


\section{Introduction}
The rising demand for on-device large language models (LLMs) has imposed a significant computational burden on mobile devices, especially in developing countries with limited hardware and infrastructure. This challenge has spurred research interest in the efficient deployment of mobile edge computing (MEC) to assist resource-constrained environments. Although conventional approaches--on-device and cloud-based inference--offer baseline support, both face critical limitations. On-device LLM inference is compute- and memory-intensive; for example, running a Llama2-7B model requires approximately 28 GB of memory, far beyond the capacity of most mobile devices \cite{laskaridismobile}. Even high-end smartphones often suffer from battery drain and degraded quality of experience (QoE). Meanwhile, cloud-based solutions like ChatGPT and Gemini introduce latency, bandwidth costs, and privacy concerns.

To address these challenges, mobile-edge collaborative inference has emerged as a promising solution. By distributing computation between mobile devices and nearby edge servers, this approach enables real-time LLM inference by integrating edge computing, model optimization, and mobile hardware. Existing collaborative methods \cite{bin2024coacto, huang2022real} focus primarily on partitioning deep neural networks (DNNs) and offloading subsequent computation stages to edge servers. For example, \cite{bin2024coacto} proposes spatially tiling the computation graph to enable independent execution among tiles, which works well for convolutional neural networks. However, such methods are ill-suited for autoregressive language models, where sequential token dependencies make partitioning inherently challenging.

To mitigate the inefficiencies of autoregressive generation, speculative decoding \cite{leviathan2023fast, chen2023accelerating} employs a lightweight \textit{draft} model on the mobile device and a more powerful \textit{target} model on the edge server. The draft model generates draft tokens with minimal computational cost, while the target model verifies and refines them in batches. This significantly reduces computational overhead while preserving inference accuracy. However, speculative decoding still faces latency issues, mainly due to asynchronous executions. As sequence lengths grow, a \textit{mutual waiting problem} arises-- the target model remains idle while awaiting tokens from the draft model, and the draft model similarly stops during target-side verification. This inefficiency hinders the overall performance gain and suggests opportunities for further optimization. 

To address this, \cite{liu2024parallel} proposes parallel speculative decoding (PSD), which overlaps the drafting and verifying phases to reduce overall processing time. However, PSD introduces additional communication overhead,  limiting its feasibility in mobile-edge collaborative inference. Motivated by these challenges, this paper highlights the need for intelligent user association and resource allocation (UARA) strategies to fully exploit the benefits of speculative decoding in MEC systems. Furthermore, we incorporate device-level constraints--such as remaining battery life--into UARA decisions for sustainable and efficient operations, given the heavy reliance of speculative decoding on mobile-side computation.

Building on these insights, we propose a novel UARA optimization framework for efficient parallel speculative decoding. Our approach addresses user association via Two-Phase Matching-based Association (TMA) \cite{wang2025uav} and applies a Multi-Agent Soft Actor-Critic (MASAC) \cite{haarnoja2018soft} network to optimize resource allocation. To the best of our knowledge, this is the first work to jointly optimize UARA problem for speculative decoding in MEC environments.

Our key contributions are summarized as follows:
\begin{itemize}[leftmargin=0.5cm] 
    \item \textbf{Unified UARA strategy for speculative decoding:} We propose the first unified framework that jointly optimizes UARA problem to enable low-latency LLM inference via parallel speculative decoding.
    \item \textbf{Joint optimization problem formulation:} We identify synchronization between mobile-side computation and uplink communication as critical to fully utilizing parallel speculative decoding. To this end, we formulate a mixed-integer, non-convex optimization problem, which is addressed using our proposed MADRL-based solution.
    \item \textbf{Realistic simulation using Sionna:} We use the Sionna \cite{hoydis2022sionna} simulator to conduct realistic experiments based on advanced ray tracing and geo-located wireless channels. This allows us to closely mimic real-world MEC systems with dynamic user mobility and varying channel conditions.
    \item \textbf{Performance evaluation and analysis:} We implement our UARA scheme on Sionna-based setups with a diverse set of mobile devices and edge servers. Numerical results show up to 28.0\% and an average of 23.7\% reduction in end-to-end latency without compromising inference accuracy, thus validating the practicality of our approach in realistic MEC settings. 
\end{itemize}

\section{Preliminaries}
\begin{figure*}[!t]
\centering
\includegraphics[width=\linewidth]{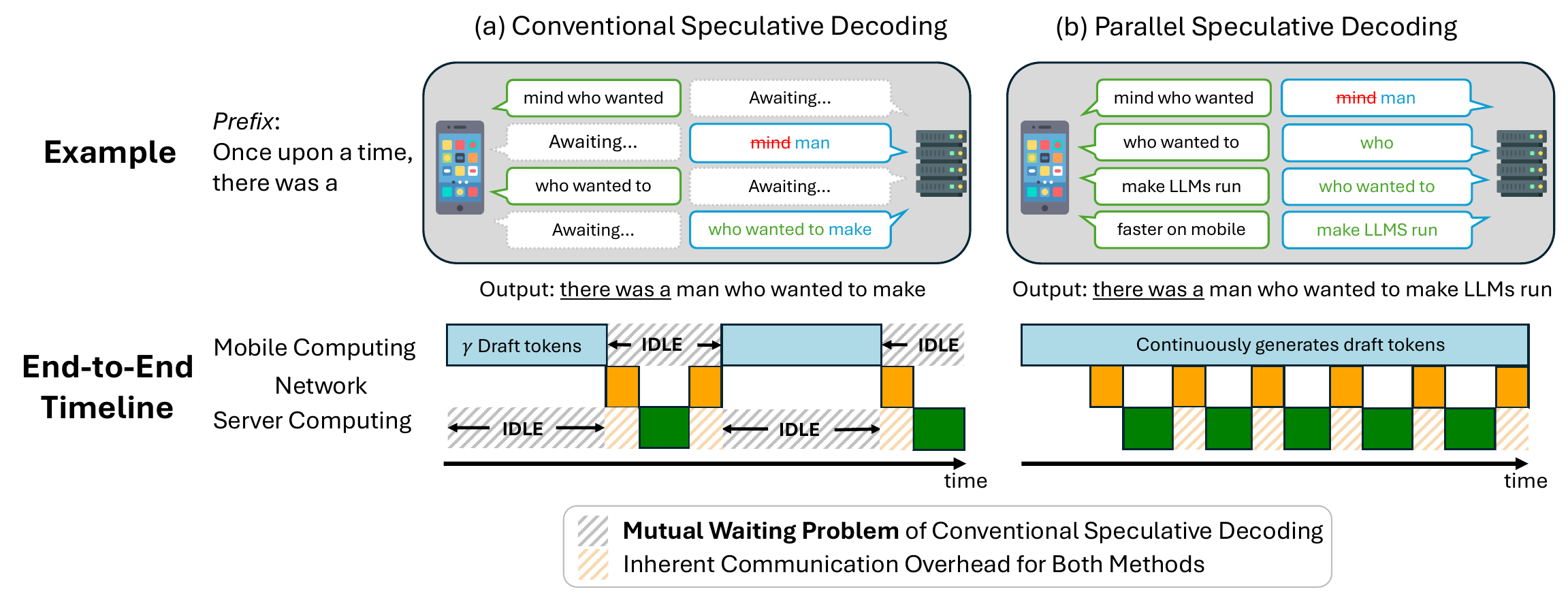}
\caption{Comparison of (a) conventional and (b) parallel speculative decoding. \textbf{(Top)} Example of token generation. In (a), the draft model waits for server-side verification and the target model waits for draft generation. In contrast, (b) allows continuous generation for both models, resulting in faster task completion. \textbf{(Bottom)} End-to-end timeline illustrating the interplay between \textcolor{SkyBlue}{mobile computation}, \textcolor{Dandelion}{network transmission} and \textcolor{Green}{server computation}. In (a), the sequential draft-then-verify process introduces prolonged idle time. In contrast, (b) reduces idle periods by enabling concurrent execution. The shaded regions highlight the \textcolor{gray}{mutual waiting problem} unique to the conventional approach and the \textcolor{orange}{communication overhead} present in both methods.}
\label{fig:1}
\end{figure*}

\subsection{Iterative LLM Inference}

The LLM inference process is inherently iterative and typically consists of two distinct phases: \textit{prompt processing (prefill)} and \textit{autoregressive generation}. These phases differ significantly in their computational and memory characteristics.

\begin{itemize}[leftmargin=0.5cm] 
    \item \textbf{Prompt Processing:} In this phase, the model processes the initial sequence \(\mathbf{x}=(x_1, ..., x_n)\) to generate the first new token \(x_{n+1}\) by computing the conditional probability \(P(x_{n+1} | x_1, ..., x_n)\). This step involves generating key-value (KV) pairs for all input tokens, which are cached for use in subsequent decoding steps. Since the entire input sequence is available upfront, the phase can be easily accelerated on modern hardware. Thus, prefill is typically compute-bound and exhibits high throughput.

    \item \textbf{Autoregressive Generation:} In the generation phase, the model produces one token at a time, with each token conditioned on all previously generated tokens. Generation continues until a stopping criterion is met, such as generating an end-of-sequence (EOS) token or reaching a specified maximum token limit. Unlike the prefill phase, this phase is primarily memory-bound since frequent memory access dominates latency. Consequently, token throughput in this phase is significantly lower, posing a major bottleneck in real-time LLM inference.
\end{itemize}

\subsection{Speculative Decoding}
Parallelizing autoregressive token generation is challenging due to its inherent sequential dependency, where each token depends on all previous ones. Speculative decoding mitigates this latency by adopting a \textit{draft-then-verify} strategy using a lightweight \textit{draft} model for fast token generation and a more accurate \textit{target} model for verification. 


In speculative decoding, the draft model \(M_q\) and the target model \(M_p\) operate with a fixed hyperparameter \(\gamma\), which specifies the draft length--the number of tokens generated by the draft model per iteration. The process begins with \(M_q\) generating \(\gamma\) draft tokens \((x_1,x_2,...,x_\gamma)\) along with their corresponding logits \((q_1,q_2,..,q_\gamma)\). The target model \(M_p\) then processes the concatenated sequence \((x, x_1,...,x_\gamma)\) and produces logits \((p_1,p_2,...,p_{\gamma+1})\). Each draft token \(x_i\) is verified with the acceptance rate \(\alpha_i\) defined as
\begin{equation}
  \alpha_i = \begin{cases}
      1 & p_i[x_i] \geq q_i[x_i] \\
      \frac{p_i[x_i]}{q_i[x_i]} & p_i[x_i] < q_i[x_i]
  \end{cases}.
\end{equation}
If a draft token \(x_i\) is rejected, it is resampled from \(p_i-q_i\); otherwise, the process proceeds to generate \(x_{\gamma +1}\)via \(p_{\gamma+1}\). This method ensures that each step generates at most \(\gamma+1\) tokens, thereby speeding up the generation process. However, we eliminate resampling and directly regenerate rejected tokens using the target model, reducing communication overhead from transmitting full logits \(q_i\). 

While most prior works focus mainly on reducing computation latency-- via compact or draft-free models \cite{zhao2024ouroboros, cai2024medusa, fu2024break} or improved alignment between draft and target distributions \cite{zhou2023distillspec, miao2024specinfer}-- these are often limited to single-device setups. Only few recent efforts have explored speculative decoding in MEC contexts \cite{hao2024hybrid, jin2024collm}. However, they often overlook the interplay between computation and communication, lacking practical and feasible strategies for real-world deployments.

\subsection{Parallel Speculative Decoding}
We introduce a parallel speculative decoding framework, inspired by recent work in \cite{liu2024parallel}, to enable synchronized execution of draft and target models. Unlike conventional speculative decoding, which alternates between draft and verify stages, PSD proposes dynamic scheduling--\textit{pre-verify} and \textit{post-verify}-- for concurrent execution. These strategies adaptively switch based on token verification outcomes, effectively overlapping computation and reducing idle time. Figure \ref{fig:1} shows the differences between conventional and parallel speculative decoding in terms of token generation behavior and execution timeline. 

In the pre-verify phase, the target model \(M_p\) computes logits \(M_p(\mathbf{x})\) while the draft model generates a batch of tokens. If the first token is rejected, the batch is discarded and a new drafting round begins immediately with the updated prefix \(\mathbf{x}+(y_1)\). This enables a fast recovery from early rejections. If accepted, the process enters the post-verify phase, where the draft model continues generating tokens without interruption and the target model concurrently verifies them. A rejection during this phase triggers a return to pre-verify phase.

By dynamically switching between these two phases, the PSD framework maintains uninterrupted utilization of both models and reduces idle time. However, we observe that this synchronization is fragile and can easily degrade when mobile-side computation and network latency are not properly aligned.

\subsection{Motivating Observations}
\begin{figure}[t]
  \centering
  \begin{subfigure}[b]{0.50\linewidth}
    \centering
    \includegraphics[width=\linewidth]{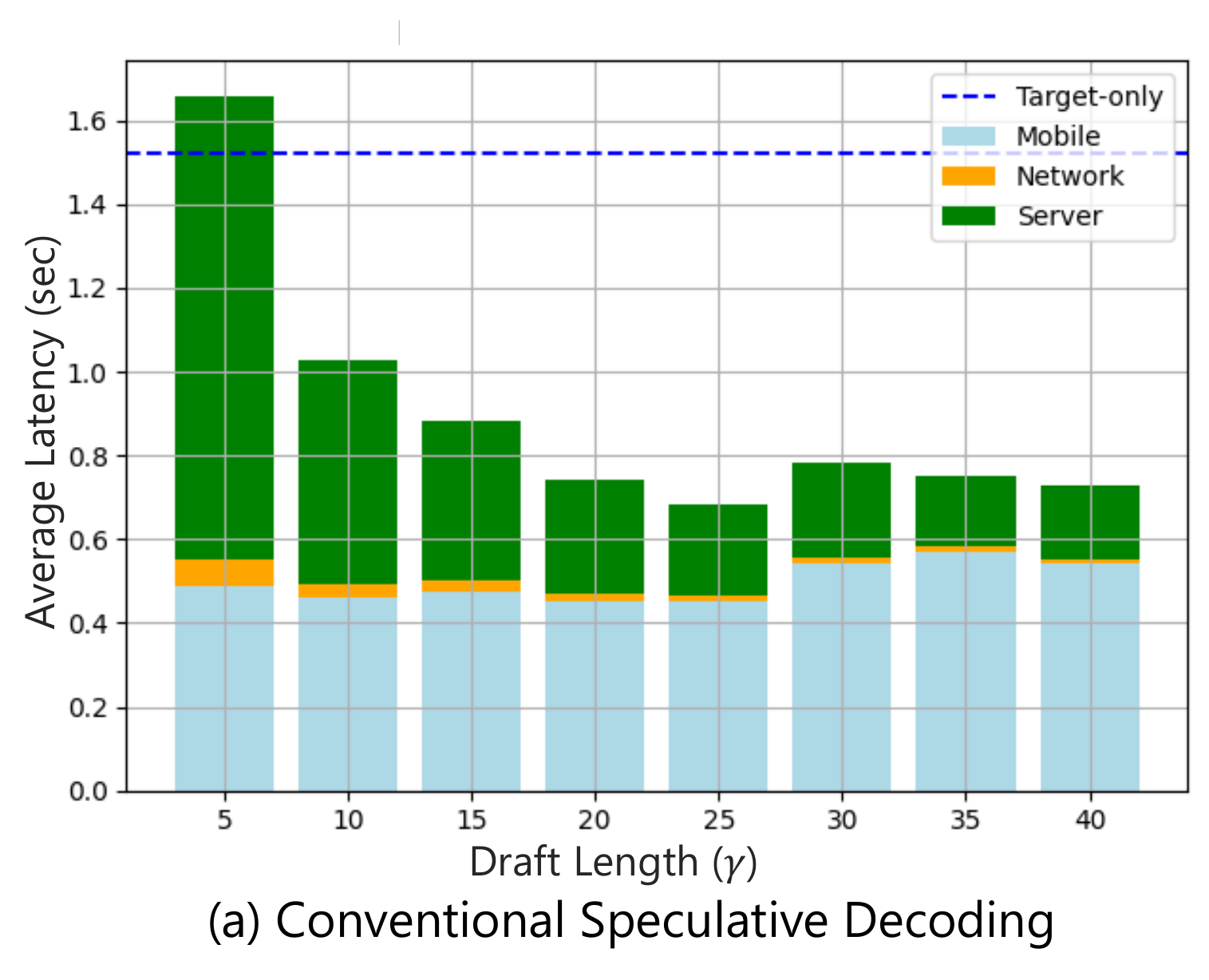}
  \end{subfigure}
  \hfill
  \begin{subfigure}[b]{0.48\linewidth}
    \centering
    \includegraphics[width=\linewidth]{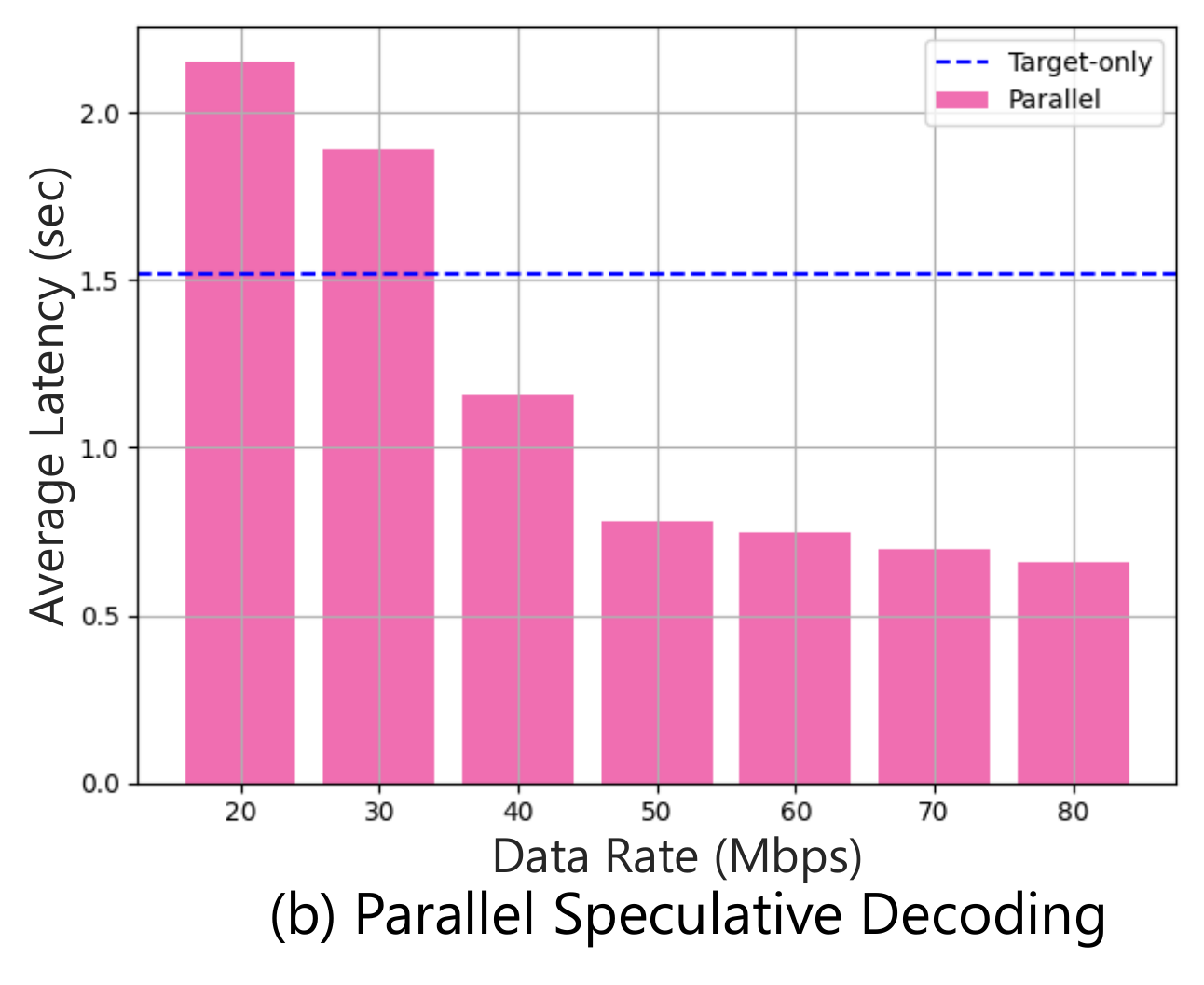}
  \end{subfigure}
  \caption{Latency of (a) conventional and (b) parallel speculative decoding using HumanEval dataset. \textbf{(Left)} Latency is broken down into \textcolor{SkyBlue}{mobile computation}, \textcolor{Dandelion}{network transmission} and \textcolor{Green}{server computation} for varying draft lengths. \textbf{(Right)} \textcolor{Rhodamine}{Total latency} is measured across different data rates.}
\label{fig:2}
\end{figure}

Figure \ref{fig:2} presents the total execution time for conventional and parallel speculative decoding under varying conditions. The key limitations of each method are summarized as follows:

\begin{itemize}
    \item \textbf{Conventional speculative decoding} lacks adaptability across diverse prompts due to its fixed draft length \(\gamma\). The optimal \(\gamma\) is \(25\) for HumanEval and \(15\) for DailyMail dataset. This motivates the adoption of parallel speculative decoding for adaptive and task-aware generation.
    \item \textbf{Parallel speculative decoding} benefits from higher bandwidth, but saturates beyond 50 Mbps, where mobile-side computation becomes the bottleneck. This underscores the need for tight synchronization between mobile computation and uplink communication.
\end{itemize}

\section{System Model and Problem Formulation}
\subsection{System Overview}
 \begin{figure}[t]
  \centering
  \includegraphics[width=\linewidth]{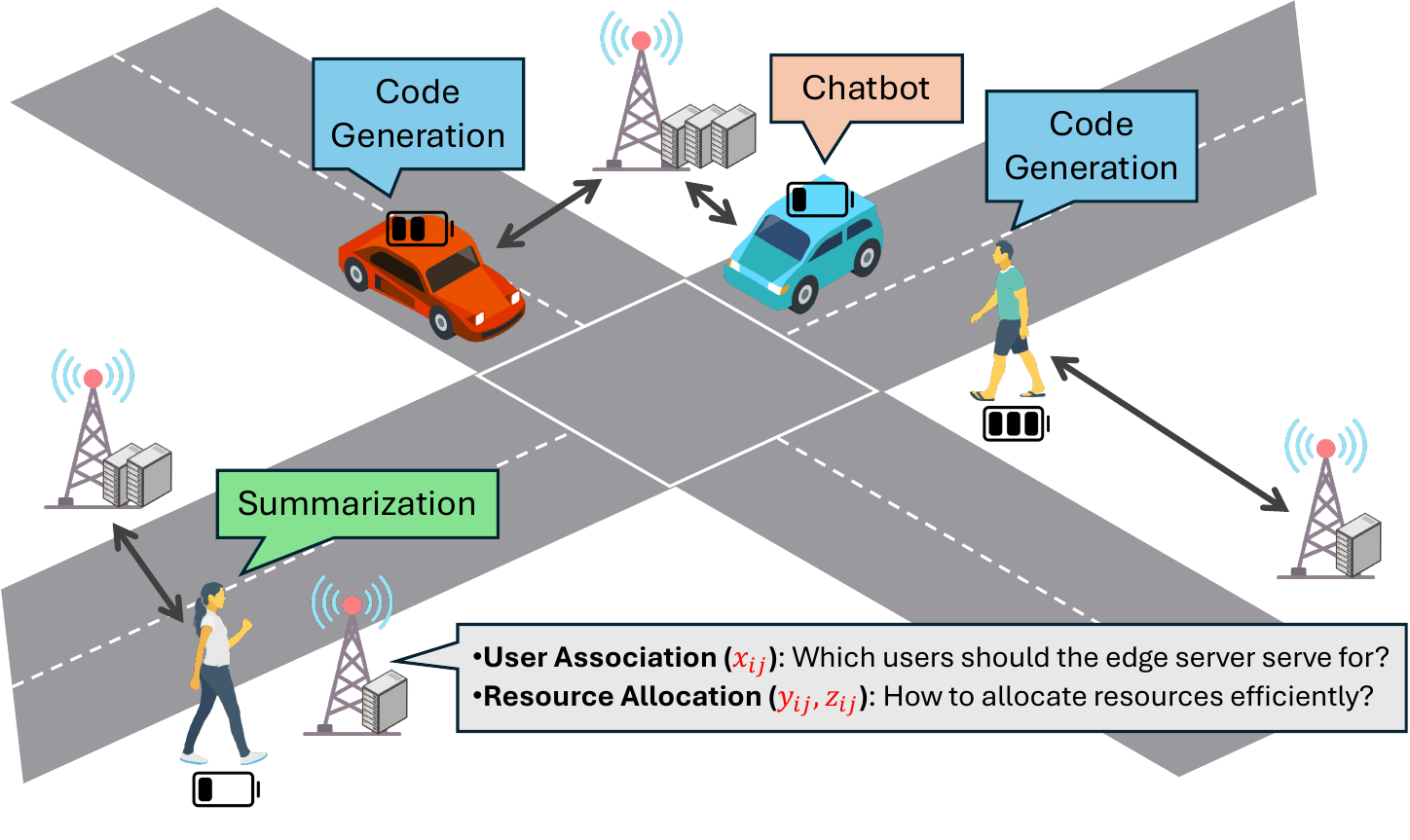}
  \caption{Overview of the proposed MEC system. Key UARA decision variables are indicated in \textcolor{red}{red}.}
  \label{fig:3}
\end{figure}

Figure \ref{fig:3} illustrates our collaborative MEC system, which consists of \(M\) mobile devices and \(E\) edge servers. The set of mobile devices is denoted by \(\mathcal{M}\), where each device may exhibit varying mobility patterns and remaining battery levels. The set of edge servers is denoted by \(\mathcal{E}\), equipped with varying computing resources. 

To capture the effect of user mobility and time-varying channel conditions, we adopt a time-slotted model that divides the offloading period into \(T\) slots with an equal duration $\tau$. 
In this quasi-static setting, channel states and device parameters are assumed constant within a slot.
Each mobile device \(m_i\) generates a task per time slot, which is represented by a tuple \(\{d_i, f_i^{MD}, f_i^{ES}\}\), where \(d_i\) is the communication load in bits, and \(f_i^{MD}\) and \(f_i^{ES}\) are the required FLOPs for mobile device and edge server, respectively. 

We then define the association of mobile device \(m_i\) to edge server \(e_j\) using a binary decision variable.
\begin{equation}
x_{ij} = 
\begin{cases}
1, & \text{if edge server } j \text{ serves mobile device } i, \\
0, & \text{otherwise}.
\end{cases}
\end{equation}
Each mobile device must offload its task to exactly one edge server, with no partial offloading allowed. Accordingly, the user association variable \(x_{ij}\) is constrained by
\begin{equation}
    \sum_{j\in \mathcal{E}} x_{ij}=1, x_{ij}\in \{0,1\}, \forall i\in\mathcal{M}, j \in \mathcal{E}
    \label{eq:3}
\end{equation}

\subsection{Communication Model}
We assume that the total network bandwidth \(W\) is equally divided among the \(E\) edge servers, such that each server \(e_j\) is allocated with \(W_j =W/E\). Each server further allocates its bandwidth \(W_j\) among the mobile devices it serves. Let \(y_{ij}\in [0,1]\) denote the fraction of \(W_j\) allocated to mobile device \(m_i\). To prevent overlapping allocations, \(y_{ij}\) is constrained by 
\begin{equation}
    \sum_{j \in \mathcal{E}} y_{ij}\leq 1, y_{ij}\in [0,1], \forall i\in \mathcal{M}, j\in \mathcal{E}.
\label{eq:4}
\end{equation}

Let \(h_{ij}\) denote the channel gain between device \(m_i\) and edge server \(e_j\). The corresponding signal-to-noise ratio (SNR) is given by \(SNR_{ij}=\frac{h_{ij}^2 P_i}{N_0}\), where \(P_i\) is the transmission power of device \(m_i\) and \(N_0\) is the noise power spectral density. Then, the data rate \(R_i\) of mobile device \(m_i\) is expressed as
\begin{equation}
    \label{eq:5}
    R_i=\sum_{j\in \mathcal{E}} x_{ij}y_{ij}W_j \log (1+SNR_{ij}).
\end{equation}

\subsection{Computing Model}
Computing latency for a task requiring \(f_i\) FLOPs is calculated by dividing the task size by the available computing capacity. For remote execution, we model the delay in two phases: a \textit{waiting phase} over \(k\) time slots of queuing delay and a \textit{computing phase} based on the server's processing power. Since edge servers share resources across tasks, we define \(z_{ij}\) as the fraction of computing capacity allocated to the task from mobile device \(m_i\). The resulting computing delays for local and remote execution are given by
\begin{equation}
\begin{cases}
D_i^{\mathrm{MD}} = \dfrac{f_i}{F_i^{\mathrm{MD}}}, & \text{(local execution)} \\
D_{ij}^{\mathrm{ES}} = k\kappa + \dfrac{f_i}{z_{ij} F_j^{\mathrm{ES}}}. & \text{(remote execution)}
\end{cases}
\end{equation}

Similar to Eq. \eqref{eq:4}, \(z_{ij}\) is constrained by
\begin{equation}
    \sum_{i \in \mathcal{M}} z_{ij}\leq 1, z_{ij}\in [0,1], \forall i\in \mathcal{M}, j\in \mathcal{E}.
    \label{eq:7}
\end{equation}

\subsection{Problem Formulation}
Effective parallel speculative decoding requires synchronization between mobile computation and uplink communication overhead. Additionally, minimizing edge-side computing latency is critical for reducing end-to-end delay. To ensure sustainable operation in resource-constrained MEC settings, we further incorporate remaining battery life of mobile devices \cite{kim2024distributed}. Based on these considerations, we formulate a multi-objective cost function that balances latency synchronization, remote computation delay, and energy efficiency.

Here, we define \(\delta^{CP}, \delta^{CM}\) as the computing and transmission energy efficiencies, respectively. The energy consumption of mobile device \(m_i\) can then be expressed as follows.
\begin{equation}
\begin{cases}
E_i^{\mathrm{CP}} = \delta^{\mathrm{CP}} f_i^{\mathrm{MD}}, & \text{(mobile computing)}\\
E_{i}^{\mathrm{CM}} = \delta^{\mathrm{CM}}\sum_{j\in \mathcal{E}}(x_{ij}y_{ij}W_j)P_i. & \text{(uplink transmission)}
\end{cases}
\end{equation}

Then, our optimization objective \(\mathcal{I}\) can be formulated as
\begin{flalign}
    \tag{$\mathcal{P}_1$}
    \min_{\mathbf{X}, \mathbf{Y}, \mathbf{Z}} \quad 
    & \sum_{i\in \mathcal{M}} \sum_{j\in \mathcal{E}} 
    x_{ij} \left( D_i^{\mathrm{MD}} + D_{ij}^{\mathrm{ES}} + \lambda \left\Vert D_i^{\mathrm{MD}} - \frac{d_i}{R_i} \right\Vert_2^2 \right) && \notag \\
    & + w \sum_{i\in \mathcal{M}} \frac{E^{\mathrm{CP}}_i + E^{\mathrm{CM}}_i}{B_i} && \notag \\
    \text{s.t.} \quad & \text{(\ref{eq:3})}, \text{(\ref{eq:4})}, \text{(\ref{eq:7})}. && \notag
    \label{problem:1}
\end{flalign}
where \(\lambda\) is a penalty for latency synchronization, and \(w\) is a unit conversion factor normalizing energy efficiency. For clarity, we define the decision variable matrices \(\mathbf{X}, \mathbf{Y}\) and \(\mathbf{Z}\) to represent user association, bandwidth allocation, and computing resource allocation, respectively. 

The primary challenge in Problem \(\mathcal{P}_1\) lies in its formulation as a mixed-integer, non-convex optimization problem. The binary variable \(\mathbf{X}\) introduces combinatorial complexity and is further coupled with the continuous variables \(\mathbf{Y}\) and \(\mathbf{Z}\), making the problem particularly difficult to solve.

\section{MADRL-based UARA Scheme}
\subsection{Two-Phase Matching-based Association (TMA)}
Two-Phase Matching-based Association (TMA) solves the user association problem through a two-phase process. In the first phase, each mobile device pre-evaluates edge servers based on channel gain to establish a favorable initial association for subsequent optimization steps. In the second phase, a matching algorithm iteratively swaps device-server pairs to minimize the objective function \(\mathcal{I}\) formulated in Problem \(\mathcal{P}_1\) until no further improvements are possible. 
Compared to traditional heuristics such as greedy search, matching-based methods offer faster convergence with lower complexity~\cite{deng2021throughput}, making them well-suited for real-time applications. The overall procedure is summarized in Alg. \ref{alg:1}. 
\begin{algorithm}[t]
    \caption{Two-phase Matching-based Association (TMA)}
    \label{alg:1}
    \begin{algorithmic}[1]
    \STATE \textbf{Initialization: } Obtain the channel gain matrix \(\mathbf{H}\) between all mobile devices and edge servers.
    \STATE \textbf{Phase 1: } Channel Gain-based Pre-evaluation
    \FORALL{\(m \in \mathcal{M}\)}
        \STATE Select the edge server \(e\) with the highest gain \(\mathbf{H}_{m,e}\).
        \STATE Set \(\mathbf{X}_{m,e}\leftarrow 1\). \\
    \ENDFOR
    \STATE \textbf{Phase 2: } Iterative Swap-based Optimization
    \WHILE {swap pair exists}
        \FORALL{\((m, e), (m', e')\) with \(\mathbf{X}_{m,e} = \mathbf{X}_{m',e'} = 1\)}
        \STATE Compute \(\mathcal{I}_{(m,e),(m',e')}\) and \(\mathcal{I}_{(m,e'),(m',e)}\).
        \IF{\(\mathcal{I}_{(m,e'),(m',e)} > \mathcal{I}_{(m,e),(m',e')}\)}
        \STATE Set \(\mathbf{X}_{m,e'} \leftarrow 1\) and \(\mathbf{X}_{m',e} \leftarrow 1\).\\
        \STATE Reset \(\mathbf{X}_{m,e} \leftarrow 0\), \(\mathbf{X}_{m',e'} \leftarrow 0\).
        \ENDIF
        \ENDFOR
    \ENDWHILE
    \end{algorithmic}
\end{algorithm}

\subsection{Proposed TMA-MASAC Framework}
We adopt a deep reinforcement learning (DRL) framework to address the resource allocation problem. The system state \(s(t)\) includes mobile device and edge server positions, channel gain matrix \(\mathbf{H}\), task information \(\{d_i, f_i^{MD}, f_i^{ES}\}\), remaining battery level \(B_i\) for each mobile device \(m_i\), and queuing delay \(k_j\) at each edge server \(e_j\). The action \(a(t)\) consists of uplink bandwidth allocation \(\mathbf{Y}(t)\) and edge computing resource allocation \(\mathbf{Z}(t)\). The reward \(r(t)\) is defined as the normalized objective value, dividing \(\mathcal{I}\) by a baseline obtained from random user association and uniform resource allocation.

Specifically, we use a multi-agent soft actor-critic (MASAC) network for two reasons. First, the continuous variables \(\mathbf{Y}\) and \(\mathbf{Z}\) require a model-free, policy-gradient method. Second, SAC jointly optimizes expected reward and policy entropy, improving robustness against highly dynamic MEC settings.

The proposed framework consists of a policy network \(\pi_{\theta}(a_t|s_t)\), a critic network \(Q_{\phi}(s_t,a_t)\), and a target critic network \(Q_{\bar{\phi}}(s_t,a_t)\). Introducing \(\alpha\) as the temperature coefficient of policy entropy, MASAC optimizes the following objective
\begin{equation}
    J(\pi)=\sum_{t=1}^T \mathbb{E}_{(s_t,a_t)\sim \pi_\theta}[r(s_t,a_t)-\alpha \log(\pi(\cdot|s_t))]
\end{equation}
To improve sample efficiency, we adopt offline sampling by reusing historical data. Mini-batches sampled from a replay buffer are used to update the policy and critic networks, while the target critic network is softly updated, following \(\bar{\phi_i}\leftarrow \xi \phi_i + (1-\xi)\bar{\xi_i}\), where \(\xi\) denotes the soft update coefficient.

\begin{figure}[h]
  \centering
  \includegraphics[width=0.99\linewidth]{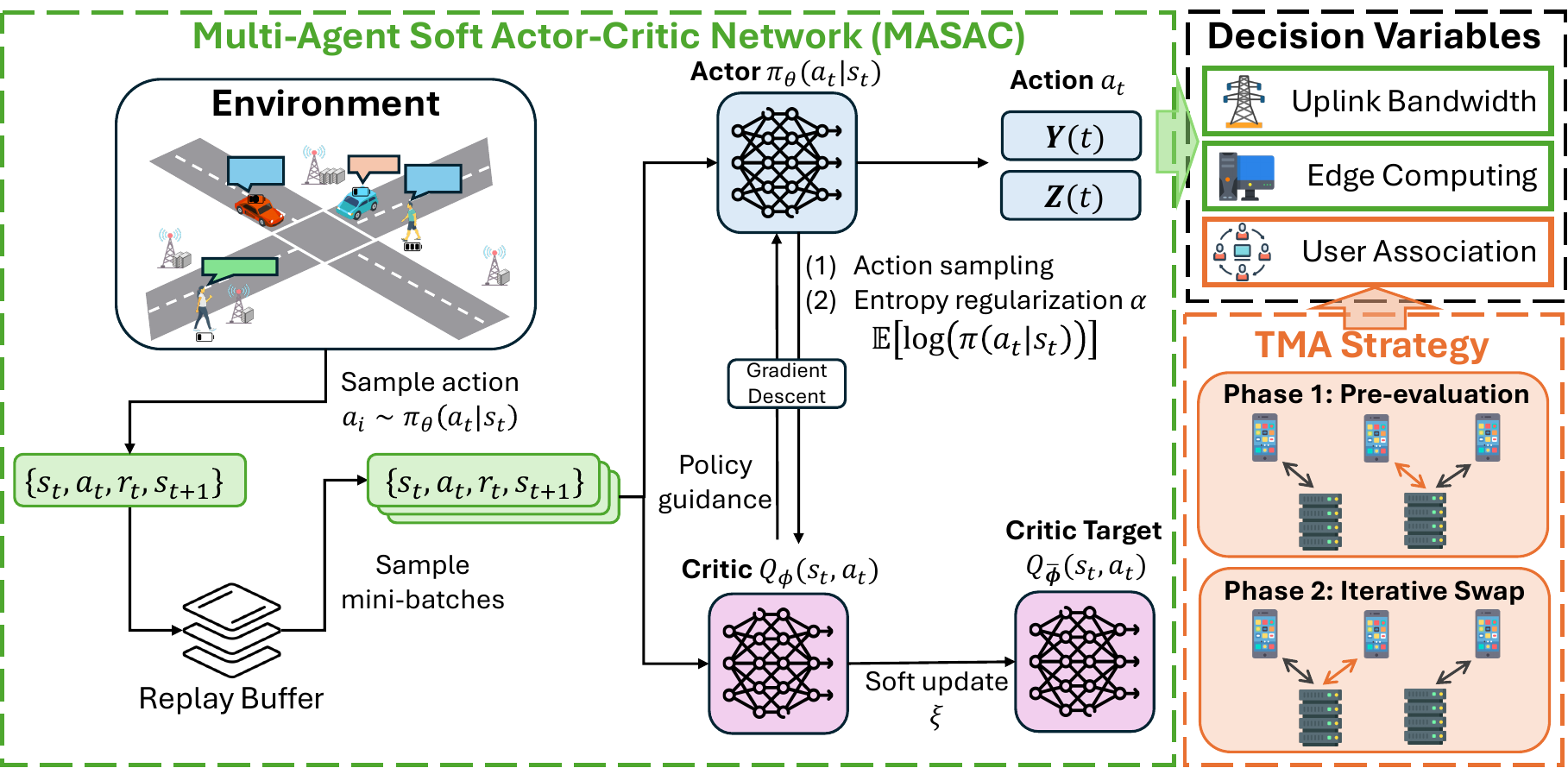}
  \caption{Illustration of the proposed TMA-MASAC.}
  \label{fig:4}
\end{figure}

To jointly solve UARA problem, the TMA-based user association is integrated into the MASAC scheme. Fig. \ref{fig:4} shows the overall framework. Specifically, at each time step, the policy network outputs resource allocation decisions, after which Alg. \ref{alg:1} is applied to determine user association \(\mathbf{X}\) based on the updated \(\mathbf{Y}\) and \(\mathbf{Z}\). 



\section{Performance Evaluation}
\subsection{Experiment Setup}
\begin{table}[t]
  \caption{MEC Simulation Configurations}
  \label{tab:1}
  \centering
  \begin{tabular}{>{\centering\arraybackslash}m{4.8cm} >{\centering\arraybackslash}m{3.2cm}}
    \toprule
    \textbf{Parameter} & \textbf{Value} \\
    \hline \hline
    Target model \(M_p\) & GPT2-XL (1.6B) \\
    Draft model \(M_q\) & GPT2 (137M) \\ \hline
    Number of mobile devices \(M\) & 30 -- 180 \\
    Number of edge servers \(E\) & 4 -- 20 \\ \hline
    Bandwidth \(W\) (MHz) & 10 \\
    Mobile transmission power \newline \(P\) (dBm) & 16 -- 24 \\
    Noise power spectral density \newline \(N_0\) (dBm/Hz) & \(-174\) \\ \hline
    Computing energy \newline coefficient \(\delta^{CP}\) & \(10^9\) \\
    Communication energy coefficient \(\delta^{CM}\) & 2.6 \\ \hline
    Latency weight \(\lambda\) & \(10^{-2}\) \\
    Energy weight \(w\) & 20 -- 100 \\
    \bottomrule
  \end{tabular}
\end{table}
\textbf{Simulation Configuration} We model the MEC system based on the 3GPP small-cell simulation guidelines \cite{3gpp2020nr}, with parameters detailed in Table \ref{tab:1}. The system includes a diverse set of mobile device and edge server, including Samsung Galaxy S23, Apple iPhone 14, Huawei Mate 60, and NVIDIA RTX 2080, 3090 and 4090. At each time slot, each device randomly executes one of three LLM-based tasks: code generation, text summarization, or chatbot conversation. Diversity in task generation with different expected sequence lengths enables an effective evaluation of the system’s adaptability to heterogeneous LLM workloads. Furthermore, to emulate realistic wireless conditions, we adopt the Sionna simulator, which offers a high-fidelity ray-tracing engine and stochastic channel models. Given the locations of mobile devices and edge servers, Sionna computes channel gain \(\mathbf{H}\), path loss, multi-path fading, and mobility dynamics. 

\textbf{Benchmarks} We evaluate the proposed method against three baselines. 
\begin{itemize}[leftmargin=0.5cm] 
    \item \textbf{Random}: Each mobile device randomly selects an edge server with uniform probability.
    \item \textbf{Max SINR}~\cite{liu2022deep}: Each mobile device associates with the edge server offering the highest signal-to-interference-plus-noise ratio (SINR).
    \item \textbf{Max Compute}~\cite{ma2021dynamic}: Each mobile device associates with the edge server offering the highest computational capacity. 
\end{itemize}

\subsection{Performance Evaluation}
\begin{figure}[h]
  \centering
  \begin{subfigure}[b]{0.49\linewidth}
    \centering
    \includegraphics[width=\linewidth]{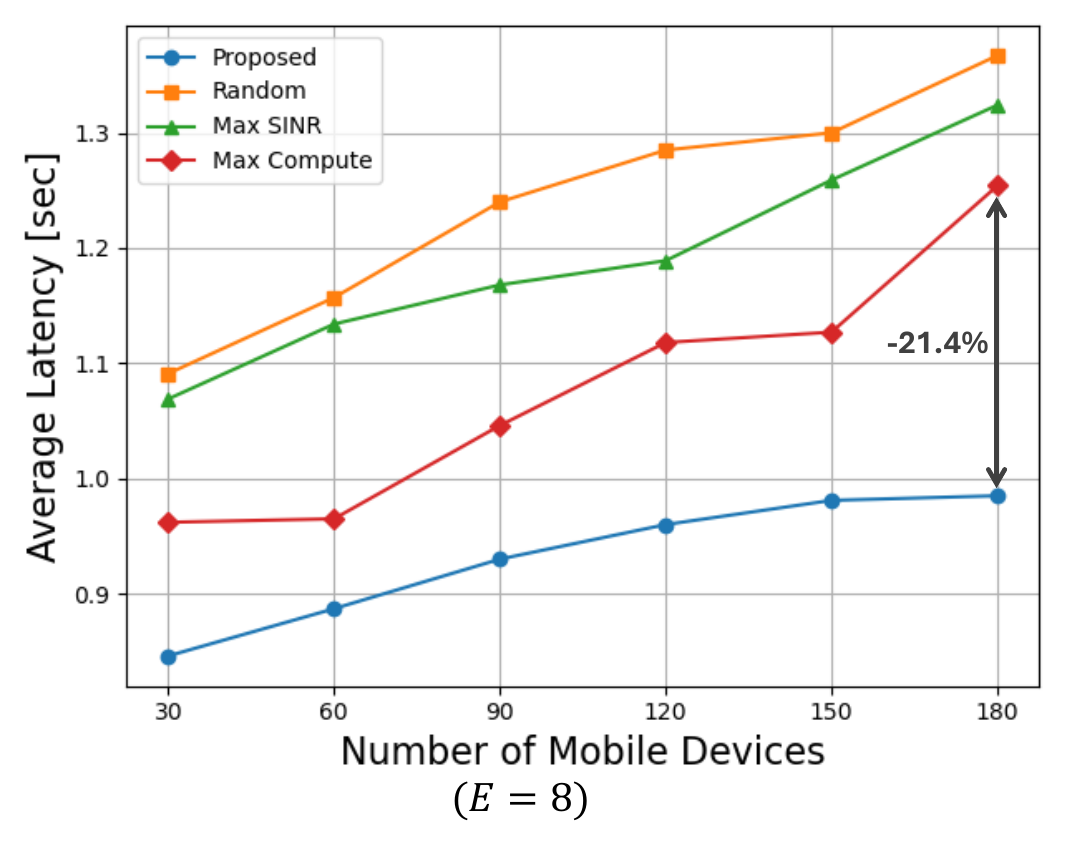}
  \end{subfigure}
  \hfill
  \begin{subfigure}[b]{0.49\linewidth}
    \centering
    \includegraphics[width=\linewidth]{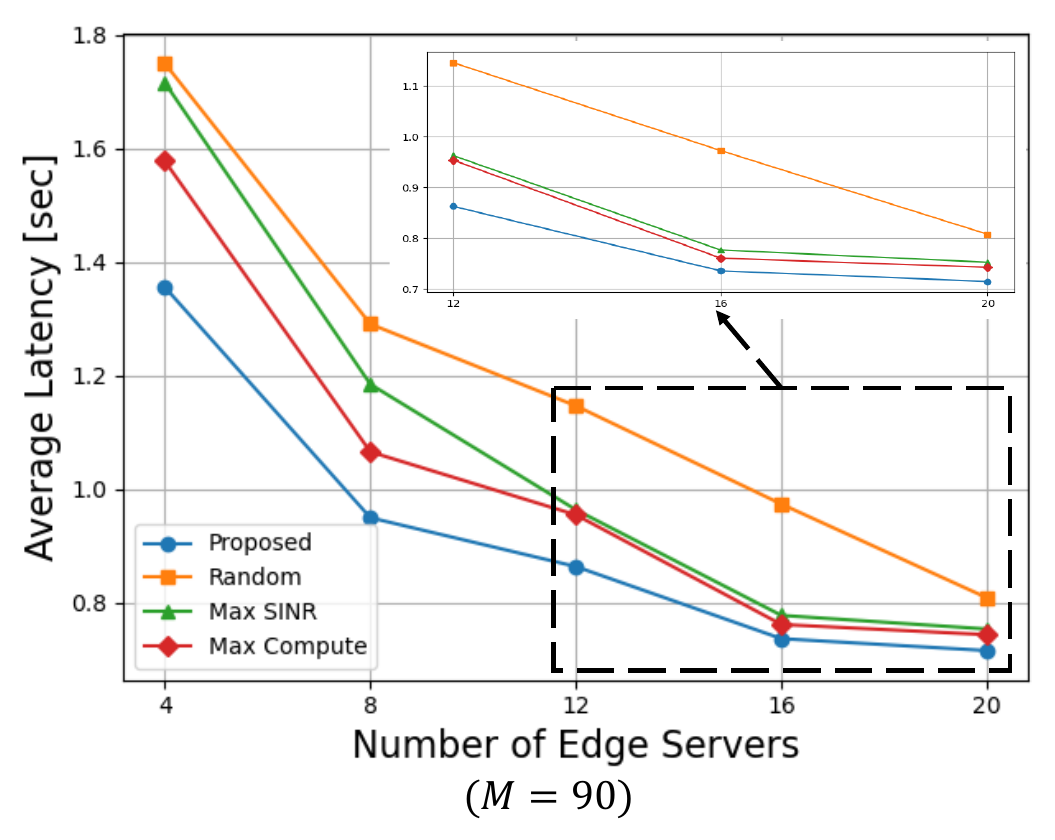}
  \end{subfigure}
  \caption{Average latency under different numbers of \textbf{(left)} mobile devices and \textbf{(right)} edge servers.}
\label{fig:5}
\end{figure}
Figure \ref{fig:5} compares the latency performance of the proposed framework against multiple baseline strategies under varying numbers of mobile devices and edge servers. As the network scales up, our method consistently outperforms reference schemes, achieving up to 28.0\% latency reduction, with an average improvement of 23.7\% across all evaluated settings. This notable gain highlights the advantage of our joint UARA strategy, which enables collaborative LLM inference with efficient resource utilization and minimizes mutual waiting delays even in dense and dynamically changing MEC environments.

\begin{figure}[h]
  \centering
  \begin{subfigure}[b]{0.49\linewidth}
    \centering
    \includegraphics[width=\linewidth]{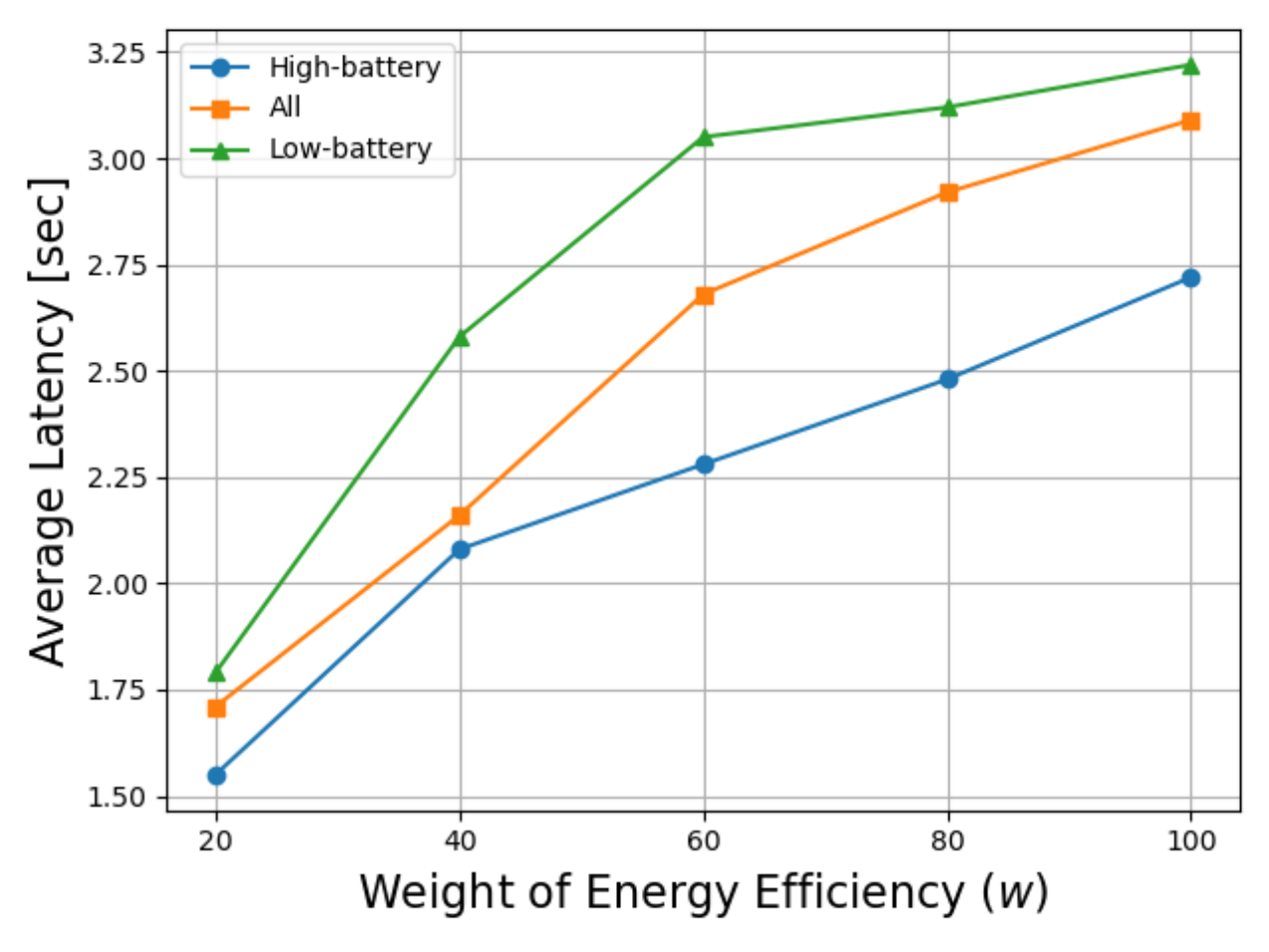}
  \end{subfigure}
  \hfill
  \begin{subfigure}[b]{0.49\linewidth}
    \centering
    \includegraphics[width=\linewidth]{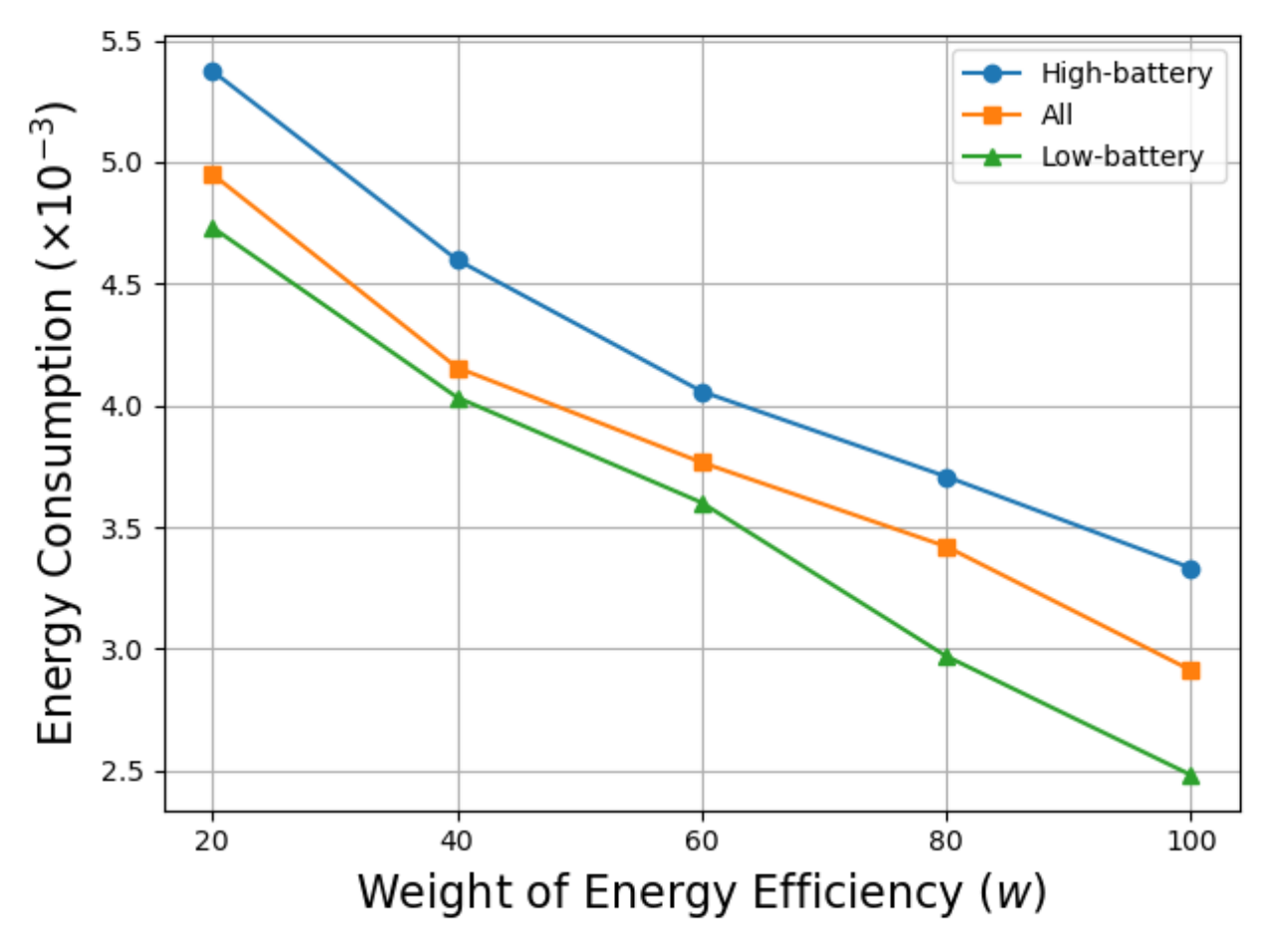}
  \end{subfigure}
  \caption{Impact of energy-efficiency weight \(w\) on \textbf{(left)} average latency and \textbf{(right)} energy consumption for mobile devices with high \((B_i \geq 0.8)\) and low \((B_i \leq 0.2)\) battery levels.}
\label{fig:6}
\end{figure}

Figure \ref{fig:6} further analyzes how the energy‐efficiency weight $w$ influences both average latency and energy consumption across devices with different battery levels. As $w$ increases, a clear tradeoff emerges: energy consumption steadily decreases while latency gradually grows, demonstrating that the system correctly prioritizes energy savings when requested. This impact is especially pronounced for low‐battery users, where the framework shifts toward energy‐conservative scheduling, preventing premature device shutdown while sustaining inference capability. These results emphasize the importance of carefully tuning w to balance latency and energy efficiency for different deployments---e.g., delay‐sensitive applications versus long‐duration, battery---constrained usage scenarios. 

\section{Conclusion}
In this paper, we introduce a framework for efficient LLM inference in mobile edge computing (MEC) systems by jointly optimizing the user–association and resource–allocation (UARA) problem to support parallel speculative decoding. To achieve this, we propose the TMA-MASAC framework, which coordinates mobile computation, wireless transmission, and server-side execution through matching algorithm and deep reinforcement learning.
Using the high‐fidelity Sionna simulator, our evaluation demonstrates that the proposed method achieves up to 28.0\% latency reduction across various network conditions, consistently outperforming existing baselines. The results confirm that integrating parallel speculative decoding with intelligent network control can accelerate LLM inference with minimal energy cost, thereby realizing scalable and energy-efficient AI deployment in dense MEC environments.

\bibliographystyle{IEEEtran}
\bibliography{{IEEEabrv,bibtex}}

\end{document}